%% file: arxiv.tex
\documentclass[twoside]{article}

\usepackage[accepted]{aistats2025}
\usepackage{hyperref}       
\usepackage{url}            
\usepackage{booktabs}       
\usepackage{amsfonts}       
\usepackage{nicefrac}       
\usepackage{microtype}      
\usepackage{xcolor}         
\usepackage{changes}
\usepackage{amssymb}
\usepackage{enumitem}
\usepackage{graphicx}
\usepackage{color}
\usepackage{subcaption}
\usepackage{amsmath}
\usepackage{amssymb}
\usepackage{amsthm}
\usepackage{graphicx}
\usepackage{tabularx}
\usepackage{cleveref}
\usepackage{fancyvrb}
\usepackage{comment}
\usepackage{algorithm}
\usepackage[noend]{algpseudocode}
\usepackage{float}
\usepackage{wrapfig}
\usepackage{listings}
\usepackage{natbib}
\usepackage{mathtools}
\usepackage{booktabs}
\usepackage{siunitx}
\usepackage{svg}
\usepackage{dblfnote}
\usepackage{lipsum}
\usepackage{dsfont}
\usepackage{multirow}
\usepackage{placeins}
\usepackage{xr} 
\input{math_commands}

\newcommand{\peter}[1]{\textcolor{red}{\textbf{Peter:} #1}}
\newcommand*{\seiyun}[1]{{\textcolor{violet}{\textbf{Seiyun:} #1}}}
\newcommand*{\iscomment}[1]{{\textcolor{blue}{\textbf{IS --- #1}}}}

\begin{document}

%

%

\twocolumn[

\aistatstitle{Dynamic DBSCAN with Euler Tour Sequences}

\aistatsauthor{ Seiyun Shin$^{1}$ \And Ilan Shomorony$^{1}$ \And  Peter Macgregor$^{2}$ }

\aistatsaddress{\hspace{4em} $^{1}$University of Illinois Urbana-Champaign 
\And $^{2}$University of St Andrews } ]

\begin{abstract}
We propose a fast and dynamic algorithm for Density-Based Spatial Clustering of Applications with Noise (DBSCAN) that efficiently supports online updates.
Traditional DBSCAN algorithms, designed for batch processing, become computationally expensive when applied to dynamic datasets, particularly in large-scale applications where data continuously evolves.
To address this challenge, our algorithm leverages the Euler Tour Trees data structure, enabling dynamic clustering updates without the need to reprocess the entire dataset.
This approach preserves a near-optimal accuracy in density estimation, as achieved by the state-of-the-art static DBSCAN method~\citep{esfandiari2021almost}. 
Our method achieves an improved time complexity of $O(d \log^3(n) + \log^4(n))$ for every
data point insertion and deletion, where $n$ and $d$ denote the total number of updates and the data dimension, respectively.
Empirical studies also demonstrate significant speedups over conventional DBSCANs in real-time clustering of dynamic datasets, while maintaining comparable or superior clustering quality.
\end{abstract}

\input{sections_aistats/1_intro}

\input{sections_aistats/2_preliminaries}

\input{sections_aistats/3_dynamic_DBSCAN}

\input{sections_aistats/4_theory}

\input{sections_aistats/5_experiments}

\input{sections_aistats/6_discussion}

\acknowledgments{The work of Ilan Shomorony was supported in part by the National Science Foundation (NSF) under grant CCF-2046991.}

\bibliography{references}

\appendix
\onecolumn
\input{sections_aistats/appendix}

\end{document}

%% file: math_commands.tex
\usepackage{amsmath,amsfonts,amssymb,bm}

\def\1{\mathds{1}}

\def\eps{{\epsilon}}

\DeclareMathAlphabet{\mathsfit}{\encodingdefault}{\sfdefault}{m}{sl}
\SetMathAlphabet{\mathsfit}{bold}{\encodingdefault}{\sfdefault}{bx}{n}

\newcommand{\E}{\mathbb{E}}

\newcommand{\R}{\mathbb{R}}

\newcommand*{\Bern}{\mathrm{Bern}}

\usepackage{algorithm, multicol, algorithmicx}
\usepackage{algpseudocode}
\algrenewcommand\algorithmicrequire{\textbf{Input: }}
\algrenewcommand\algorithmicensure{\textbf{Output: }}
\algnewcommand\algorithmicinput{\textbf{Input:}}

\newcommand{\allnotation}[1]{#1}

\newtheorem{theorem}{Theorem}

\newtheorem{lemma}{Lemma}
\newtheorem{claim}{Claim}
\newtheorem{remark}{Remark}
\newtheorem{definition}{Definition}
\newtheorem{corollary}{Corollary}

\newtheorem{assumption}{Assumption}

\newcommand{\abs}[1]{\left\lvert#1\right\rvert}
\newcommand{\calB}{\mathcal{B}}
\newcommand{\calF}{\mathcal{F}}

\newcommand{\calX}{\mathcal{X}}
\newcommand{\HB}{\textsf{bucket}}

\newcommand{\dist}{\textsf{dist}}
\newcommand{\idx}{\textsf{idx}}
\renewcommand{\vec}[1]{{\allnotation{\bm{#1}}}}

\newcommand{\vecx}{\vec{x}}
\newcommand{\vecy}{\vec{y}}

\renewcommand{\epsilon}{\varepsilon}

\newcommand{\Z}{\mathbb{Z}}

\newcommand{\union}{\cup}
\newcommand{\intersect}{\cap}
\newcommand{\cardinality}[1]{\abs{#1}}

\newcommand{\blue}[1]{\textcolor{blue}{#1}}



%% file: sections_aistats/1_intro.tex
\section{INTRODUCTION} \label{sec:intro}
Density-based clustering is a fundamental problem in data science, with machine learning applications spanning computer vision~\citep{shen2016real}, and medical imaging~\citep{tran2012density, baselice2015dbscan}.
The key idea behind density-based clustering is to group points in space by identifying connected regions with high data density while separating them from sparser areas.
One of the most widely used methods in this category is Density-Based Spatial Clustering of Applications with Noise (DBSCAN), introduced by~\citet{ester1996density}.
DBSCAN estimates the density of each point by counting the number of points within its neighborhood and classifies those with densities above a threshold as core points.
A neighborhood graph of core points is then constructed, with clusters formed based on the connected components.
This method has been successfully adopted and integrated into numerous data mining tools~\citep{hall2009weka, pedregosa2011scikit, r2013r, schubert2015framework}.

Despite DBSCAN's success across many domains, it faces two major challenges: (1) its quadratic time complexity and (2) the requirement to process the entire dataset upfront.
These limitations make it impractical for large-scale datasets that evolve \emph{dynamically} over time.
Static algorithms with polynomial-time complexity are no longer practical for such massive and evolving data.


To address the first challenge, several efficient and scalable algorithms have been proposed~\citep{chen2005geometric, gunawan2013faster, gan2017hardness, de2019faster, jang2019dbscan++}.
However, as the dimensionality $d$ of the data increases, these algorithms still exhibit a running time similar to $O(n^2)$, where $n$ denotes the number of data points.
A notable improvement is DBSCAN\texttt{++}~\citep{jang2019dbscan++}, which achieves faster clustering while maintaining the accuracy of the original DBSCAN under $\beta$-regularity assumption (to be specified later).
Building on the work from~\citet{rinaldo2010generalized} that reveals the relationship between connected components of a neighborhood graph and the density level sets,
DBSCAN\texttt{++} has been shown to be near-optimal for $\lambda$-density level set estimation with respect to \emph{Hausdorff distance}, with a time complexity of $O(n^{2-\tfrac{2\beta}{2\beta+d}})$.
However, even this method struggles to avoid quadratic time complexity for high-dimensional datasets.

More recently, a near-linear time algorithm with $O(dn\log^2 (n))$ complexity was proposed by~\citet{esfandiari2021almost} which makes use of locality sensitive hash funcions, and achieves a near-optimal Hausdorff distance for density level sets.
While this method offers significant improvements due to the decoupled dependency on the number of data points $n$ and the data dimension $d$ in its runtime guarantee, the algorithm remains designed primarily for static, batch-processing tasks.
Consequently it becomes computationally expensive for dynamic datasets where data points are continuously added or removed.
In particular, using this static algorithm for even a single point update requires reprocessing the entire dataset, meaning it incurs the same time complexity as clustering all $n$ data points, making it inefficient for large-scale applications.


Our work addresses both the computational and dynamic challenges by introducing a fast, dynamic DBSCAN algorithm that efficiently supports online updates.
By leveraging the \emph{Euler Tour Sequence} data structure for dynamic forests~\citep{henzinger1995randomized, tseng2019batch}, our algorithm enables dynamic clustering updates without the need to reprocess all data points, thereby significantly reducing computational overhead.
This approach preserves a near-optimal accuracy in density estimation as achieved by the state-of-the-art static DBSCAN method~\citep{esfandiari2021almost}, with a time complexity of $O(d \log^3 (n) + \log^4 (n))$ for every data point insertion and deletion.
Here $n$ denotes the maximum number of data points at any time.

Through extensive empirical studies, we demonstrate that our algorithm provides substantial speed improvements over the state-of-the-art static DBSCAN, as well as the conventional DBSCAN method, particularly for real-time clustering in dynamic datasets. Not only does it deliver superior computational efficiency, but it also maintains or improves clustering quality.

%% file: sections_aistats/2_preliminaries.tex
\section{PRELIMINARIES}{\label{sec:preliminaries}}

Consider a dynamic setting where the dataset $X \subseteq \R^d$ evolves over time as new data points are added or existing ones are removed.
We assume that each data point is drawn independently and identically distributed (i.i.d.) from a distribution $\calF$ over $\R^d$.
The goal is to \emph{dynamically} partition the evolving dataset into distinct clusters based on the density of the data points.
\subsection{DBSCAN} \label{subsec:dbscan}
The foundational DBSCAN algorithm~\citep{ester1996density} builds on the idea of identifying high-density regions in a static dataset to form clusters.
The algorithm first estimates points that belong to density level set, where points in dense regions are referred to as core points.
These core points serve as an approximation of the density level set.
The algorithm then identifies connected components of these dense regions by constructing a graph where core points within distance $\eps$ are connected.
The final clusters are the connected components of this graph, with non-core points assigned to nearby dense regions and remaining points treated as noise or outliers.
According to~\Cref{alg:dbscan}, a point $\vec{x}$ is classified as a core point if there are at least $k$ points within its $\eps$-radius neighborhood $\calB(\vec{x}, \eps)$
The neighborhood is determined by a distance metric, which is defined in Euclidean space as follows:
\begin{definition} \label{def:dist}
    Given $\vec{x}, \vec{y} \in \mathbb{R}^d$, and a set $C \subseteq \mathbb{R}^d$, define $\dist(\vec{x}, \vec{y}) := \|\vec{x} - \vec{y}\|_2 = \sqrt{\sum_{i=1}^d(x_i - y_i)^2}$; also define a distance between $\vec{x}$ and $C$ to be $\dist(\vec{x}, C) := \inf_{\vec{y} \in C} \|\vec{x} - \vec{y}\|_2$.
    For $\eps > 0$, define $\calB(\vec{x}, \eps): = \{\vec{y} \in \calX: \dist(\vec{x}, \vec{y}) \leq \eps\}$ and $\calB(C, \eps) := \{\vec{y} \in \calX: \dist(\vec{y}, C) \leq \eps\}$, respectively.
\end{definition}

\begin{algorithm}[t]
    \caption{\textsc{DBSCAN}$(k, \varepsilon, X)$} \label{alg:dbscan}
    \begin{algorithmic}[1]
       \State Initialise set of core points $C = \emptyset$
       \State Initialise empty graph $G$
       \For{$\vec{x} \in X$}
       \State $C \gets C \cup \{\vec{x}\}$, if $\lvert \{\vec{y} \in X: \dist(\vec{x}, \vec{y}) \leq \varepsilon \}\rvert \geq k$
       \EndFor
       \State Construct a graph $G = (V, E)$, where $V \gets X$
       \For{$\vec{c} \in C$}
       \State Add edge $e = (\vec{c}, \vec{x}) \in E$, $\forall \vec{x} \in X \intersect \calB(\vec{c}, \eps)$
       \EndFor
       \State Return connected components of $G$
    \end{algorithmic}
\end{algorithm}
To establish theoretical guarantees, subsequent works \citep{jiang2017density, jang2019dbscan++} assume the existence of an underlying, yet \emph{unknown}, density function $f: \calX \to \R_{\geq 0}$, where $\calX \subseteq \R^d$ denotes the support of $\calF$.
The density of any given data point $\vec{x}$ is then defined as $f(\vec{x})$.
In particular, $\lambda$-density level set (or simply $\lambda$-level set) is defined as:
\begin{definition}[Density Level Set] \label{def:lambda}
    We define the $\lambda$-level set of $f$ as:
    $L_f(\lambda) := \left\{ \vec{x} \in \calX: f(\vec{x}) \geq \lambda \right\}$.
    Here $\lambda$ characterises the density level.
\end{definition}
The algorithm is given $\lambda$ as input, but does not have access to $f$.
Building on the work from~\citet{rinaldo2010generalized}, which demonstrates that partitioning via a neighborhood graph can effectively approximate the clustering of level sets,~\citet{jiang2017density, jang2019dbscan++} provide comprehensive studies on how to choose $(k, \eps)$ 
to estimate the density level set for a given threshold $\lambda$, and provide bounds on the Hausdorff distance error.
In practical situations, where $\lambda$ is not explicitly provided, $(k, \eps)$ are treated as hyperparameters to optimise the clustering performance.

Following this framework, our proposed algorithm will also use $(k, \epsilon)$ as hyperparameters, with theoretical guarantees based on the $\lambda$-level set.
The relationship between $(k, \eps)$ and the density level $\lambda$ will be explicitly discussed in~\Cref{sec:theory}.

\subsection{Dynamic Forests} \label{subsec:dynamic_forest}
A dynamic forest is a data structure which provides the following operations:
\begin{itemize}[noitemsep, topsep=0.2pt]
    \item \textsc{add}$(\vecx)$: add a new node to the forest.
    \item \textsc{link}$(\vecx, \vecy)$: connect $\vecx$ and $\vecy$, if they are in different trees.
    \item \textsc{cut}$(\vecx, \vecy)$: remove any edge between $\vecx$ and $\vecy$.
    \item \textsc{root}$(\vecx)$: return the root of the tree having $\vecx$.
\end{itemize}
\citet{henzinger1995randomized} proposed the use of Euler Tours Trees in order to implement this data structure, with a running time of $O(\log (n))$ for the \textsc{link}, \textsc{cut}, and \textsc{root} operations (also we refer to~\citet{demaine2005advanced}).
Instead of storing the represented forest directly, this approach stores the Euler Tour Sequence of each tree in the forest.
By viewing the tree as a directed graph, where each edge is represented by two directed edges, the Euler Tour Sequence is 
the sequence of nodes visited when performing an Euler tour of the tree beginning with the root.

\citet{henzinger1995randomized} demonstrated that the dynamic forest operations can be efficiently implemented by storing the Euler Tour Sequences in a balanced binary tree data structure, with appropriate joining and splitting when edges are added or removed from the forest.

More recently,~\citet{tseng2019batch} proposed another scheme based on the same idea, in which the Euler Tour Sequences are instead represented using a skip list data structure.
This approach preserves the same asymptotic guarantees while simplifying the implementation considerably, and it is this technique that we employ for our proposed DBSCAN algorithm.

%% file: sections_aistats/3_dynamic_DBSCAN.tex
\section{DYNAMIC DBSCAN ALGORITHM}

\begin{algorithm}
    \caption{\textsc{DynamicDBSCAN}} \label{alg:main}
    \begin{algorithmic}[1]
    \Procedure{Initialise$(k, t, \varepsilon)$}{}
       \State Initialise hash functions $\{h_i\}_{i=1}^t$, dynamic forest $G$, and set of core points $C = \emptyset$.
    \EndProcedure
    \Procedure{AddPoint$(\vecx)$}{}
        \State G.\textsc{add}($\vecx$)
        \State $C' = \emptyset$ \Comment{New core points}
        \For{$i \in [t]$}
            \State Compute $h_i(\vecx)$ and add $\vecx$ to $\HB_i(\vecx)$
            \If{$\cardinality{\HB_i(\vecx)} > k$}
                \State $C' \gets C' \union \{\vecx\}$
            \ElsIf{$\cardinality{\HB_i(\vecx)} = k$}
                \State $C' \gets C' \union (\HB_i(\vecx) \setminus C)$
            \EndIf
        \EndFor
        \State $C \gets C \cup C'$
        \For{$\vec{c} \in C'$}
            \State \textsc{LinkCorePoint}$(\vec{c})$
        \EndFor
        \If{$C' = \emptyset$}
            \State \textsc{LinkNonCorePoint}$(\vecx)$
        \EndIf
    \EndProcedure
    \Procedure{DeletePoint}{$\vecx$}
        \If{$\vecx$ is a core point}
            \State $C' = \emptyset$ \Comment{New non-core points}
            \For{$i \in [t]$}
                \If{$\cardinality{\HB_i(\vecx)} = k$}
                    \State Add each $\vecy \in \HB_i(\vecx)$ to $C'$ if $\cardinality{\HB_j(\vecy)} < k$ for all $j \neq i$.
                \EndIf
            \EndFor
            \State $C \gets C \setminus C'$
            \For{$\vec{c} \in C'$}
                \State \textsc{UnlinkCorePoint}$(\vec{c})$
                \State \textsc{LinkNonCorePoint}$(\vec{c})$
            \EndFor
        \EndIf
        \State Remove $\vecx$ from $G$, $C$, and all hash tables.
    \EndProcedure
    \Procedure{LinkCorePoint}{$\vecx$}
        \State Cut any edge incident to $\vec{c}$ in $G$.
        \For{$i \in [t]$}
            \State $\vec{c}_1 \gets \max\{{\vec{c} \in \HB_i(\vec{x}): \idx(\vec{c}) < \idx(\vec{x})}\}$
            \State $\vec{c}_2 \gets \min\{{\vec{c} \in \HB_i(\vec{x}): \idx(\vec{c}) > \idx(\vec{x})}\}$
            Here $\max(\cdot)$ and $\min(\cdot)$ refer to returning the point with the highest and lowest index, respectively.
            \State G.\textsc{cut}$(\vec{c}_1, \vec{c}_2)$
            \State G.\textsc{link}$(\vec{c}_1, \vecx)$
            \State G.\textsc{link}$(\vecx, \vec{c}_2)$
        \EndFor
    \EndProcedure
    \Procedure{UnlinkCorePoint}{$\vecx$}
        \For{$i \in [t]$}
            \State $\vec{c}_1 \gets \max\{{\vec{c} \in \HB_i(\vec{x}): \idx(\vec{c}) < \idx(\vec{x})}\}$
            \State $\vec{c}_2 \gets \min\{{\vec{c} \in \HB_i(\vec{x}): \idx(\vec{c}) > \idx(\vec{x})}\}$
            \State G.\textsc{cut}$(\vec{c}_1, \vecx)$
            \State G.\textsc{cut}$(\vecx, \vec{c}_2)$
            \State G.\textsc{link}$(\vec{c}_1, \vec{c}_2)$
        \EndFor
        \State Re-link any non-core points attached to $\vecx$.
    \EndProcedure
    \Procedure{LinkNonCorePoint}{$\vecx$}
        \State Link $\vecx$ with \textbf{one} core point in $\HB_i(\vecx)$, for some $i$.
    \EndProcedure
    \Procedure{GetCluster}{$\vecx$}
        \State Return G.\textsc{root}($\vecx$)
    \EndProcedure
    \end{algorithmic}
\end{algorithm}

Our algorithm builds on the nearly-linear time algorithm for DBSCAN given by~\citet{esfandiari2021almost}.
In contrast to the standard DBSCAN described in~\Cref{alg:dbscan}, they observe that locality-sensitive hash functions can be used to efficiently find close data points without an expensive linear scan of the dataset.
We will use hash functions defined as follows:
\begin{definition}[Hash Function] \label{def:hash}
    For a point $\vec{x} \in \R^d$ and parameter $\varepsilon$, we define a hash function 
    $h(\vec{x}) := \lfloor \tfrac{\vec{x} + \eta \cdot \1_d}{2\eps} \rfloor$,
    where the floor function is applied entry-wise to the vector,
    $\eta$ is drawn independently from the uniform distribution over $[0, 2 \varepsilon]$,
    and $\1_d$ is the $d$-dimensional all-ones vector.
\end{definition}
With this definition,~\citet{esfandiari2021almost} give the following guarantee for the recovery of close points.
\begin{lemma}[\citep{esfandiari2021almost}] \label{lem:hash}
    Given $\eps > 0$, the following holds for any two points $\vec{x}, \vec{y} \in \mathbb{R}^d$ and for a hash function $h$:
    \begin{enumerate}[nolistsep, topsep=0.2pt]
        \item $\Pr[h(\vec{x}) = h(\vec{y})] \geq 1 - \tfrac{\|\vec{x} - \vec{y}\|_1}{2\epsilon}$;
        \item $h(\vec{x}) = h(\vec{y}) \implies \|\vec{x} - \vec{y}\|_\infty \leq 2\epsilon$.
    \end{enumerate}
\end{lemma}
Using this fact that close points are likely to have the same hash value, we create $t$ independent hash functions $h_1, \ldots h_t$ and define the set of core points $C$ for our DBSCAN algorithm as follows.
\begin{definition}[Core Points]
    For a dataset $X := \{\vec{x}_i\}_{i=1}^n$, the set of core points $C \subseteq X$ is
    $$C \triangleq \{ \vec{x}_i \in X: \exists j \in [t] \ \mbox{such that}\ \cardinality{\HB_j(\vec{x}_i)} \geq k \},$$
    where $\HB_j(\vec{x}_i) := \{ \vec{x}_\ell \in X: h_j(\vecx_\ell) = h_j(\vec{x}_i)\}$.
\end{definition}
We then construct a graph $H = (V, E)$, where each vertex $\vec{v}_i$ corresponds to the data point $\vec{x}_i$,
and an edge $(\vec{v}_i, \vec{v}_j) \in E$ exists between any pair of core points $\vec{x}_i, \vec{x}_j \in C$, if $\vec{x}_i$ and $\vec{x}_j$ collide under any of the hash functions $h_1, \ldots h_t$.
\citet{esfandiari2021almost} show that with appropriate choices of the parameters $\epsilon$, $t$, and $k$, the connected components of the graph $H$ correspond to the connected components of the $\lambda$-density level set of the data.

\begin{figure}[t]
    \centering
    \includegraphics[width=0.65\linewidth]{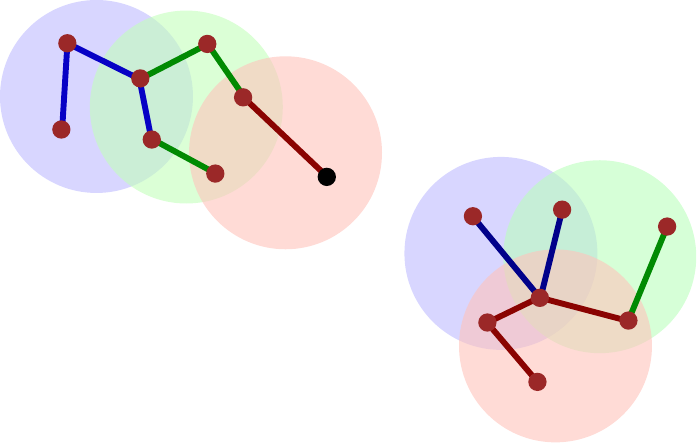}
    \caption{Illustration of a graph constructed by the \textsc{DynamicDBSCAN} algorithm: Red points represent core points, and each shaded region corresponds to a separate hash bucket.
    Edge colors match the hash bucket they belong to.
    Within each hash bucket, a path is added on the core points unless adding an edge would introduce a cycle into the graph.}
    \label{fig:dydbscan}
\end{figure}

For our dynamic application, however, maintaining the connected components of $H$ as points are added and removed from the dataset is quite challenging.
This difficulty arises primarily due to the fact that $H$ is a dense graph with no inherent structure.
In order to overcome this, we will maintain a \emph{spanning forest} of $H$ (as illustrated in~\Cref{fig:dydbscan}) throughout the dynamic updates, and guarantee that the maximum degree of any vertex in the spanning forest is at most $O(\log (n))$.
By using an Euler tour sequence data structure to maintain the dynamic spanning forest, the update time of our dynamic data structure is bounded by $O(\log^2 (n))$.

Our \textsc{DynamicDBSCAN} data structure is formally described in~\Cref{alg:main}.
In each of the \textsc{AddPoint}$(\vecx)$ and \textsc{RemovePoint}$(\vecx)$ methods, we first compute the change to the set of core points resulting from adding (resp.\ removing) $\vecx$.
Then, we update the connections for any core points added (resp.\ removed).
Within each hash bucket, we always maintain that the core points are connected in a path structure according to the indices of the data points.
This ensures that the degree of every core point in the spanning forest is bounded by $O(t)$, where $t$ is the number of hash functions, and we set it to be $O(\log (n))$.
Non-core points are each connected to at most one core point with which they collide in any hash function, ensuring that each non-core point has a degree of at most 1 in the spanning forest.

Finally, our data structure always supports a \textsc{GetCluster}$(\vecx)$ query method, which returns a unique identifier of the cluster containing $\vecx$.
This corresponds to returning the identifier of the tree containing $\vecx$ in the spanning forest.
This can be obtained through a call to the \textsc{root} operation of the dynamic forest data structure,
with a time complexity of $O(\log (n))$.

%% file: sections_aistats/4_theory.tex
\section{THEORETICAL RESULTS} \label{sec:theory}

\subsection{Time Complexity} \label{subsec:runtime}
\begin{theorem} \label{thm:runtime}
    Let $X \subseteq \R^d$ be a set of $n$ data points updated through data point insertions and deletions.
    The running time of the procedures in~\Cref{alg:main} is as follows:
    \begin{enumerate}[nolistsep, topsep=0.2pt]
        \item \textsc{Initialise}$(k, t, \varepsilon)$ runs in $O(td)$ time.
        \item \textsc{AddPoint}$(\vecx)$ and \textsc{DeletePoint}$(\vecx)$ run in $O(t^2 k (d + \log(n))$ time.
        \item \textsc{GetCluster}$(\vecx)$ runs in $O(\log(n))$ time.
    \end{enumerate}
    By setting $t = O(\log(n))$ and $k = O(\log(n))$, our data structure has update time $O(d \log^3(n) + \log^4(n))$.
\end{theorem}
\begin{proof}
    The running time of the \textsc{Initialise} procedure is dominated by the initialisation of the $t$ hash functions, each of which requires $O(d)$ time to create.
    Hence, the overall time complexity for \textsc{Initialise} is $O(td)$.

    To establish the second claim,
    we first show that the running time of the \textsc{LinkCorePoint}, \textsc{UnlinkCorePoint} and \textsc{LinkNonCorePoint} operations are all bounded by $O(t \cdot (d + \log (n)))$.
    These follow from the two facts:
    (1) computing the hash values for $\vecx$ takes \(O(td)\) time;
    and (2) the remaining operations, such as the \textsc{cut} and \textsc{link} operations, are efficiently handled using the Euler Tour Sequence data structure, such that each operation takes $O(\log (n))$ time per hash functions, and searching for core points $\vec{c}_1$ and $\vec{c}_2$ within the hash buckets is optimised using a balanced binary tree.
    Furthermore, observe that in the \textsc{AddPoint} and \textsc{DeletePoint} methods, at most $O(t k)$ calls are made, as the number of affected core points $\cardinality{C'}$ is bounded by $t \cdot k$.
    Consequently, these methods make at most $O(t k)$ calls to the \textsc{LinkCorePoint}, \textsc{UnlinkCorePoint}, and \textsc{LinkNonCorePoint} operations, resulting in a total running time of $O(t^2 \cdot k \cdot (d + \log (n)))$.

    Finally, the \textsc{GetCluster} method involves a single call to the \textsc{root} operation on the Euler tour dynamic forest, having the time complexity of $O(\log (n))$.
\end{proof}
\begin{remark}[Comparison to~\citet{esfandiari2021almost}]
    By leveraging the Euler tour sequence data structure, our algorithm efficiently handles dynamic updates without needing to reprocess the entire dataset, achieving a time complexity of $O(d \log^3(n) + \log^4(n))$.

    In contrast, the use of the the static algorithm proposed by \citet{esfandiari2021almost} requires reconfiguration of the core and non-core point sets due to the update, as well as graph reconstruction.
    This process is essentially equivalent to re-running the algorithm from scratch with $n+1$ data points, resulting in a time complexity of $O(tdn\log (n)) = O(dn\log^2 (n))$ when $t = O(\log (n))$.

\end{remark}

\begin{remark}[Memory Complexity]
    The total memory usage of Algorithm~\ref{alg:main} is $O((n+d)\log (n))$. This is due to the fact that (1) we use $t = O(\log (n))$ hash functions, requiring $O(d\log (n))$ space to store the hash functions themselves; (2) $O(n\log (n))$ space is required to store the data in the hash buckets; and lastly (3) we note that the space to store the spanning forest is $O(n)$.
\end{remark}

\subsection{Correctness of Dynamic Updates}

In this subsection, we prove that after each call to \textsc{AddPoint} or \textsc{DeletePoint}, our algorithm consistently correctly maintains that the subgraph of $G$, corresponding to the core points, as a spanning forest $H = (V_H, E_H)$.
In this forest, each vertex $\vec{v}_i$ corresponds to a core point $\vecx_i$, and 
$(\vec{v}_i, \vec{v}_j) \in E_H$ if and only if there exists some $\ell \in [t]$ such that $h_\ell(\vecx_i) = h_\ell(\vecx_j)$.
\begin{theorem} \label{thm:correct}
    After every call to \textsc{AddPoint} or \textsc{DeletePoint}, $G[C]$ (i.e., the subgraph of $G$ induced by the core points) remains a spanning forest of $H$.
\end{theorem}
Notice that since $G[C]$ is a spanning forest of $H$, and non-core points in $G$ have degree at most 1, this immediately implies that the connected components of $G$ correspond directly to the connected components of $H$.
Also, since $H$ is invariant to the order in which points are added and removed, and the clustering behaviour of \textsc{DynamicDBSCAN} is based only on connected components, it suffices to analyse $H$ in order to obtain theoretical guarantees on the clustering quality.
\begin{proof}[Proof of~\Cref{thm:correct}]
    We proceed by induction.
    Assume that prior to any call to \textsc{AddPoint} or \textsc{DeletePoint}, $G[C]$ is a spanning forest of $H$.
    Then one can readily see that when \textsc{AddPoint}$(\vecx)$ (resp.\ \textsc{DeletePoint}$(\vecx)$) is called, the algorithm correctly computes the set $C'$ which represents the change in the core set induced by adding (resp.\ removing) the point $\vecx$.
    
    Let $b_{i, j} := \{ \vecx \in X: h_i(\vecx) = j\}$ be a hash bucket of $h_i$.
    By the inductive hypothesis that $G[C]$ is a spanning forest of $H$ prior to the update, we see that every core point in $b_{i, j}$ is in the same connected component of $G$.
    
    For \textsc{AddPoint}, \Cref{alg:main} calls \textsc{LinkCorePoint} for each new core point in $C'$.
    Following each call to \textsc{LinkCorePoint}, the algorithm maintains that every core point in $b_{i, j}$ is in the same connected component of $G$:
    notice that in each bucket, a link between $\vec{c}_1$ and $\vec{c}_2$ may be cut, but linking $\vec{c}_1$ and $\vec{c}_2$ again with the new point $\vecx$ ensures that $\vec{c}_1$ and $\vec{c}_2$ remain in the same connected component.

    The case for \textsc{RemovePoint} is similar: the algorithm calls \textsc{UnlinkCorePoint}$(\vecx)$ for each point removed from the set of core points.
    Although the links $(\vec{c}_1, \vecx)$ and $(\vecx, \vec{c}_2)$ are cut, the algorithm ensures that $\vec{c}_1$ and $\vec{c}_2$ remain connected by calling \textsc{link}$(\vec{c}_1, \vec{c}_2)$.

    We highlight that the \textsc{link}$(\vec{x}, \vec{y})$ operation of the dynamic forest data structure adds an edge between $\vec{x}$ and $\vec{y}$ only if they are not already in the same tree.
    This ensures that the structure of $G$ remains a forest throughout the execution of the algorithm.

    Finally, since edges are only added between core points that collide in some hash bucket, $G[C]$ is always a subgraph of $H$.
    Given that:
    \begin{itemize}[noitemsep, topsep=0.2pt]
        \item $G[C]$ is a subgraph of $H$;
        \item every pair of points connected by an edge in $H$ are in the same connected component of $G[C]$; and
        \item $G[C]$ is a forest;
    \end{itemize}
    we can conclude that $G[C]$ is a spanning forest of $H$ after every call to \textsc{AddPoint} or \textsc{RemovePoint}.
\end{proof}

\subsection{Approximation of Density Level Sets}

We are now ready to state the theoretical results for our algorithm with respect to finding density level sets.
Notably, the near-linear time DBSCAN algorithm proposed by~\citet{esfandiari2021almost} has been shown to achieve a near-optimal statistical guarantee for estimating density level sets.
In contrast to their algorithm, \textsc{DynamicDBSCAN} defines core points based on all $t$ hash functions, rather than using a dedicated hash function specifically for core point determination.
This new definition ensures that no hash bucket contains more than $k$ non-core points,
a crucial requirement to guarantee poly-logarithmic update time for the data structure.

Armed with this, we will show that despite this modification, Algorithm~\ref{alg:main}
still dynamically maintains an estimator for the $\lambda$-density level set, achieving a near-optimal performance with respect to the Hausdorff distance.




\subsubsection{Assumptions and Parameter Setup} \label{subsubsec:assump_and_parameter}
Throughout the analysis, we follow prior work~\citep{jiang2017density, jang2019dbscan++, esfandiari2021almost} and make two regularity assumptions on the density distribution of the data.
\begin{assumption} \label{assump:conti_dist}
    $f$ is continuous and has convex compact support $\mathcal{X} \subseteq \R^d$.
\end{assumption}
\begin{assumption}[$\beta$-regularity of level-sets] \label{assump:regularity}
    There exist $\beta, R_1, R_2, \lambda_c > 0$ such that
    $\forall \vec{x} \in L_f(\lambda - \lambda_c)\setminus L_f(\lambda)$, $R_1 \cdot \dist(\vec{x}, L_f(\lambda))^\beta \leq \lambda - f(\vec{x}) \leq R_2 \cdot \dist(\vec{x}, L_f(\lambda))^\beta$.
\end{assumption}
The first assumption ensures that the distribution is continuous, which is a natural requirement.
The second assumption, referred to as $\beta$-regularity, is a standard condition in level set analysis that characterises the prominence of the level set boundary.
This is captured by $\beta$ and $\lambda_c$.

For given values of $n$, $d$, $\beta$, $\lambda$, and $\lambda_c$ corresponding to our target dataset,
we choose parameters of the \textsc{DynamicDBSCAN} algorithm to satisfy
\begin{itemize}[noitemsep, topsep=0.2pt]
    \item $k \geq M_1 \cdot \left(\tfrac{\lambda}{\lambda_c}\right)^2 \cdot \omega(\log (n))$;
    \item $k \leq M_2 \cdot \lambda^{\tfrac{2\beta + 2d}{2\beta + d}}\cdot O\left((\log (n))^{\tfrac{d}{2\beta + d}}\cdot n^{\tfrac{2\beta}{2\beta + d}}\right)$;
    \item $\eps := \tfrac{1}{2}\left(\tfrac{k}{n\lambda(1-2C_{\delta, n}/\sqrt{k})}\right)^{\tfrac{1}{d}}$;
    \item $t := 100 \log \left(\tfrac{n}{\delta}\right)$;
\end{itemize}
where $C_{\delta, n} := C_0 \left(\log (t/\delta) \sqrt{d\log (n)}\right)$ and $\delta$ is a confidence parameter, ensuring that our guarantees hold with a probability of at least $1 - \delta$.
Moreover, we consider the regime where $n$ is sufficiently large with $M_1$ chosen to be sufficiently large and $M_2$ sufficiently small, both depending on $d$, and the density function $f(\cdot)$.
While we treat the dimension $d$ as a constant here, our results also hold for $d = \Theta(\log^c (n))$ for any constant $c > 0$.
Additional details can be found in the Appendix.

The following uniform convergence bound from \citet{chaudhuri2010rates} will serve as a key component in our analysis.
\begin{lemma}[Lemma 7 of~\citet{chaudhuri2010rates}] \label{lem:base}
Let $X$ be a set of $n$ i.i.d.\ samples drawn from a distribution $\mathcal{F}$ over $\mathcal{X}$.
With probability at least $1 - \tfrac{\delta}{t}$, the following conditions hold for any cube $K \subseteq \mathbb{R}^d$:
\begin{enumerate}[nolistsep, topsep=0.2pt]
    \item If $\Pr_{\vec{x} \sim \mathcal{F}} [\vec{x} \in K] \geq \tfrac{k}{n} + C_{\delta, n} \tfrac{\sqrt{k}}{n}$, then $\lvert X \cap K \rvert \geq k$.
    \item If $\Pr_{\vec{x} \sim \mathcal{F}} [\vec{x} \in K] < \tfrac{k}{n} - C_{\delta, n} \tfrac{\sqrt{k}}{n}$, then $\lvert X \cap K \rvert < k$.
    \item If $\Pr_{\vec{x} \sim \mathcal{F}} [\vec{x} \in K] \geq C_{\delta, n} \tfrac{\sqrt{d \log n}}{n}$, then $\lvert X \cap K \rvert \geq 1$. 
\end{enumerate}
\end{lemma}

\begin{table*}[t]
    \centering
    \begin{tabular}{ccccc}
        \toprule
         Name & $n$ & $d$ & Clusters & Reference \\
         \midrule
         Letter & 20000 & 16 & 26 & \cite{letter} \\
         MNIST & 70000 & 20 & 10 & \cite{mnist} \\
         Fashion-MNIST & 70000 & 20 & 10 & \cite{fashion-mnist} \\
         Blobs & 200000 & 10 & 10 & Synthetic Data \\
         KDDCup99 & 494000 & 20 & 23 & \cite{kddcup99} \\
         Covertype & 581012 & 54 & 7 & \cite{covtype_dataset} \\
         \bottomrule
    \end{tabular}
    \caption{Dataset Information}
    \label{tab:datasets}
\end{table*}

\subsubsection{Density Level Set Estimation}

We now establish that the connected components generated by our algorithm provide an accurate estimation of the target density level set.
To this end, we first demonstrate that any sampled point located too far from the desired density level set will not be included in the core point set.
Furthermore, we show that for any point within the desired density level set, there exists a nearby point that will be included in the core set.
We note that this analysis follows that of \citet{esfandiari2021almost} quite closely although we must carefully take account of the difference in our definition of core points.
These findings are formalised in the following two lemmas.

\begin{lemma} \label{lem:non-core}
    For any point $\vec{x} \in X$, if $\dist(\vec{x}, L_f(\lambda)) \geq 2 \left( \tfrac{\lambda}{R_1} \cdot 10 \tfrac{C_{\delta, n}}{\sqrt{k}} \right)^{1/\beta}$, then $\vec{x}$ will not be added to the core point set $C$, with probability at least $1 - \delta$.
\end{lemma}

\begin{proof}
    Let $\vec{x} \in X$ be such that $\dist(\vec{x}, L_f(\lambda)) \geq 2 \left( \tfrac{\lambda}{R_1} \cdot 10 \tfrac{C_{\delta, n}}{\sqrt{k}} \right)^{1/\beta}$ and consider any point $\vec{y} \in X$ satisfying $h_i(\vec{y}) = h_i(\vec{x})$ for some $i \in [t]$.
    By the second claim in~\Cref{lem:hash},
    \(
    \| \vec{x} - \vec{y} \|_2 \leq \sqrt{d} \| \vec{x} - \vec{y} \|_\infty \leq 2 \sqrt{d} \epsilon.
    \)
    Hence, the distance from $\vec{y}$ to the level set $L_f(\lambda)$ can be bounded as follows:
    \(
    \dist(\vec{y}, L_f(\lambda)) \geq \dist(\vec{x}, L_f(\lambda)) - 2 \sqrt{d} \epsilon.
    \)
    Since $\tfrac{C_{\delta, n}}{\sqrt{k}} = \left(\tfrac{1}{M_2}\right)^{\tfrac{1}{2\beta}}\cdot \omega\left((\log (n)/n)^{\tfrac{1}{2\beta + d}}\right)$ and $\eps = M_2^{\tfrac{1}{d}}\cdot O\left((\log (n)/n)^{\tfrac{1}{2\beta + d}}\right)$, (detailed derivations are provided in Appendix~\ref{app:lem3details}),
    it follows that
    \(
    \dist(\vec{y}, L_f(\lambda)) \geq \tfrac{\dist(\vec{x}, L_f(\lambda))}{2}
    \)
    for sufficiently small $M_2$.
    By~\Cref{assump:regularity}, we have
    \(
    f(\vec{y}) \leq \lambda - R_1 \cdot \dist(\vec{y}, L_f(\lambda))^\beta \leq \lambda - R_1 \cdot \left( \tfrac{\dist(\vec{x}, L_f(\lambda))}{2} \right)^\beta \leq \lambda - 10\lambda \tfrac{C_{\delta, n}}{\sqrt{k}}.
    \)
    
    Now consider the condition for $\vec{x}$ to be added to the core set. For $\vec{x}$ to be a core point, there must be at least $k$ points within its hash bucket;
    however, as
    \(
    \Pr[\vec{x} \in K] = \int_{\calX} f(\vec{y})\cdot \1_{[h_i(\vec{y}) = h_i(\vec{x})]} d\vec{y} \leq (2\eps)^d \left(\lambda - 10\lambda \tfrac{C_{\delta, n}}{\sqrt{k}}\right) \leq 
    k\cdot \tfrac{\lambda - 10\lambda C_{\delta, n}/\sqrt{k}}{n\lambda(1-2C_{\delta, n}/\sqrt{k})} \leq \tfrac{k}{n} - 8C_{\delta, n}\tfrac{\sqrt{k}}{n},
    \)
    by~\Cref{lem:base}, with probability at least $1-\tfrac{\delta}{t}$, $\lvert X \intersect K \rvert < k$.
    Applying the union bound for all $\{h_i\}_{i=1}^t$, we conclude that
    $\vec{x}$ is not included in $C$, with probability at least $1 - \delta$.
\end{proof}

\begin{lemma} \label{lem:core}
    With probability at least $1 - \delta$, any point $\vec{x} \in X$ satisfying $\dist(\vec{x}, L_f(\lambda)) \leq \tfrac{1}{2}\left( \tfrac{\lambda}{R_2} \cdot \tfrac{C_{\delta, n}}{\sqrt{k}} \right)^{1/\beta}$ will be added to $C$.
\end{lemma}

\begin{proof}
    Let $\vec{x} \in X$ such that $\dist(\vec{x}, L_f(\lambda)) \leq \tfrac{1}{2}\left( \tfrac{\lambda}{R_2} \cdot \tfrac{C_{\delta, n}}{\sqrt{k}} \right)^{1/\beta}$ and consider any point $\vec{y} \in X$ satisfying $h_i(\vec{y}) = h_i(\vec{x})$ for some $i \in [t]$. By~\Cref{lem:hash}, we have
    \(
    \| \vec{x} - \vec{y} \|_2 \leq 2 \sqrt{d} \epsilon.
    \)
    Hence,
    \(
    \dist(\vec{y}, L_f(\lambda)) \leq \dist(\vec{x}, L_f(\lambda)) + 2 \sqrt{d} \epsilon \leq \left( \tfrac{\lambda}{R_2} \cdot \tfrac{C_{\delta, n}}{\sqrt{k}} \right)^{1/\beta},
    \)
    where the last inequality follows from the same order-wise comparison provided in the proof of~\Cref{lem:non-core}.
    \Cref{assump:regularity} then implies
    \(
    f(\vec{y}) \geq \lambda - \lambda\tfrac{C_{\delta, n}}{\sqrt{k}},
    \)
    from which we have
    \(
    \Pr[\vec{x} \in K] = \int_{\calX} f(\vec{y})\cdot \1_{[h_i(\vec{y}) = h_i(\vec{x})]} d\vec{y} \geq (2\eps)^d \left(\lambda - \lambda \tfrac{C_{\delta, n}}{\sqrt{k}}\right) \geq 
    k\cdot \tfrac{\lambda - \lambda C_{\delta, n}/\sqrt{k}}{n\lambda(1-2C_{\delta, n}/\sqrt{k})} \geq \tfrac{k}{n} + C_{\delta, n}\tfrac{\sqrt{k}}{n},
    \)
    implying from~\Cref{lem:base} that $\lvert X \intersect K \rvert \geq k$, with probability at least $1 - \delta$.
    Hence, $\vec{x}$ will be added to the core point set $C$.
\end{proof}

\begin{lemma} \label{lem:close_core}
    For any point $\vec{x} \in L_f(\lambda)$, with probability at least $1 - \delta$, there exists a core point $\vec{y} \in C$ such that
    \(
    \|\vec{x} - \vec{y}\|_2 \leq \tfrac{\varepsilon}{2} \leq \tfrac{1}{2} \left( \tfrac{\lambda}{R_2} \cdot \tfrac{C_{\delta, n}}{\sqrt{k}} \right)^{1/\beta}.
    \)
\end{lemma}
\begin{proof}
    Let $r_0 = \tfrac{1}{2} \left( \tfrac{2C_{\delta,n}\sqrt{d\log (n)}}{n \lambda} \right)^{1/d}$.
    Then we obtain
    \(
    \int_{\mathcal{X}} f(\vec{z}) \cdot \1_{[\| \vec{z} - \vec{x} \|_\infty \leq r_0]} d\vec{z}
    \geq (2r_0)^d (\lambda - R_2 \cdot r_0^\beta)
    \geq (2r_0)^d \cdot \tfrac{\lambda}{2}
    = C_{\delta,n}\tfrac{\sqrt{d\log (n)}}{n},
    \)
    where the second inequality follows for sufficiently large $n$.
    By~\Cref{lem:base}, there exists a point $\vec{y} \in X$ such that $\| \vec{x} - \vec{y} \|_\infty \leq r_0$.
    Comparing the order with respect to $n$ (similar to the one in the proof of~\Cref{lem:non-core}), we see that $r_0 \leq \tfrac{\varepsilon}{2\sqrt{d}}$.
    We therefore conclude that
    \(
    \|\vec{x} - \vec{y}\|_2 \leq \tfrac{\varepsilon}{2} \leq \tfrac{1}{2} \left( \tfrac{\lambda}{R_2} \cdot \tfrac{C_{\delta, n}}{\sqrt{k}} \right)^{1/
    \beta},
    \)
    implying from~\Cref{lem:core} that $\vec{y}$ is in $C$.
\end{proof}

Now, we bound the Hausdorff error when using the core points returned by~\Cref{alg:main} to estimate the desired density level set.
\begin{definition}[Hausdorff Distance] \label{def:Haus_dist}
    Given two sets $C$ and $C'$,
    $\dist_{\textsf{Haus}}(C, C') := \max\{\sup_{\vec{x} \in C} \dist(\vec{x}, C'), \linebreak \sup_{\vec{x}' \in C'} \dist(\vec{x}', C)\}$.
\end{definition}

\begin{theorem} \label{thm:cc_dist_main}
    Let $C$ be the set of core points obtained from~\Cref{alg:main}.
    With probability at least $1 - \delta$, it follows that
    \(
    \dist_{\textsf{Haus}}(C, L_f(\lambda)) \leq 2 \left( \tfrac{\lambda}{R_1} \cdot 10 \tfrac{C_{\delta,n}}{\sqrt{k}} \right)^{1/\beta}.
    \)
\end{theorem}
\begin{proof}
   First, notice that~\Cref{lem:close_core} implies that $\sup_{\vec{x} \in L_f(\lambda)} \dist(\vec{x}, C) \leq \tfrac{1}{2} \left( \tfrac{\lambda}{R_2} \cdot \tfrac{C_{\delta, n}}{\sqrt{k}} \right)^{1/\beta}$.
   In addition,~\Cref{lem:non-core} implies that $\sup_{\vec{x} \in C} \dist(\vec{x}, L_f(\lambda)) \leq 2 \left( \tfrac{\lambda}{R_1} \cdot 10 \tfrac{C_{\delta, n}}{\sqrt{k}} \right)^{1/\beta}$.
   Given the hyperparameter choice of $k$, $\dist_{\textsf{Haus}}(C, L_f(\lambda)) \rightarrow 0$, as $n$ tends to infinity.
\end{proof}

\begin{remark}
    By selecting the maximum possible value for $k$, the resulting quantity is at most $\tilde{O}\left(M_3 \cdot n^{-\tfrac{1}{2\beta + d}}\right)$, where $M_3$ is a parameter that only depends on $(d, \delta, \lambda)$, and the density function $f$.
    This matches with the lower bound established in Theorem 4 of~\citet{tsybakov1997nonparametric},
    indicating that our density level set estimation is near optimal.
\end{remark}

%% file: sections_aistats/5_experiments.tex
\section{EMPIRICAL RESULTS} \label{sec:experiments}

 \begin{table}
    \resizebox{\columnwidth}{!}{
\begin{tabular}{ccccc}
\toprule 
 & & \multicolumn{3}{c}{Algorithm} \\
 \cmidrule{3-5}
 Dataset & Metric & \textsc{DyDBSCAN} & \textsc{EMZ} & \textsc{Sklearn} \\
 \midrule 
 \multirow{3}{*}{Letter} & Time & $ 1.44{\scriptstyle \pm  0.036}$ & $ 1.51{\scriptstyle \pm  0.011}$ & $\mathbf{ 1.10}{\scriptstyle \pm  0.046}$ \\
 & ARI & $\mathbf{ 0.02}{\scriptstyle \pm  0.001}$ & $\mathbf{ 0.02}{\scriptstyle \pm  0.002}$ & $ 0.00{\scriptstyle \pm  0.000}$ \\
 & NMI & $ 0.27{\scriptstyle \pm  0.007}$ & $\mathbf{ 0.29}{\scriptstyle \pm  0.006}$ & $ 0.20{\scriptstyle \pm  0.000}$\\
 \midrule 
\multirow{3}{*}{MNIST} & Time & $ 1.64{\scriptstyle \pm  0.027}$ & $ 2.63{\scriptstyle \pm  0.018}$ & $\mathbf{ 0.99}{\scriptstyle \pm  0.004}$ \\
 & ARI & $\mathbf{ 0.02}{\scriptstyle \pm  0.001}$ & $\mathbf{ 0.02}{\scriptstyle \pm  0.002}$ & $ 0.00{\scriptstyle \pm  0.000}$ \\
 & NMI & $ 0.22{\scriptstyle \pm  0.011}$ & $\mathbf{ 0.26}{\scriptstyle \pm  0.013}$ & $ 0.20{\scriptstyle \pm  0.000}$\\
 \midrule 
\multirow{3}{*}{Fashion-MNIST} & Time & $\mathbf{ 6.49}{\scriptstyle \pm  0.159}$ & $ 20.55{\scriptstyle \pm  0.057}$ & $ 26.17{\scriptstyle \pm  0.051}$ \\
 & ARI & $\mathbf{ 0.05}{\scriptstyle \pm  0.001}$ & $\mathbf{ 0.05}{\scriptstyle \pm  0.001}$ & $ 0.00{\scriptstyle \pm  0.000}$ \\
 & NMI & $ 0.15{\scriptstyle \pm  0.003}$ & $\mathbf{ 0.26}{\scriptstyle \pm  0.001}$ & $ 0.05{\scriptstyle \pm  0.000}$\\
 \midrule 
\multirow{3}{*}{Blobs} & Time & $\mathbf{ 84.39}{\scriptstyle \pm  1.008}$ & $ 241.96{\scriptstyle \pm  2.943}$ & $ 621.43{\scriptstyle \pm  1.921}$ \\
 & ARI & $\mathbf{ 1.00}{\scriptstyle \pm  0.001}$ & $\mathbf{ 1.00}{\scriptstyle \pm  0.000}$ & $ 0.98{\scriptstyle \pm  0.000}$ \\
 & NMI & $ 0.99{\scriptstyle \pm  0.001}$ & $\mathbf{ 1.00}{\scriptstyle \pm  0.000}$ & $ 0.97{\scriptstyle \pm  0.000}$\\
 \midrule 
\multirow{3}{*}{KDDCup99} & Time & $\mathbf{ 431.81}{\scriptstyle \pm  3.975}$ & $ 6005.93{\scriptstyle \pm  22.759}$ & - \\
 & ARI & $\mathbf{ 0.91}{\scriptstyle \pm  0.001}$ & $\mathbf{ 0.91}{\scriptstyle \pm  0.000}$ & - \\
 & NMI & $\mathbf{ 0.80}{\scriptstyle \pm  0.001}$ & $\mathbf{ 0.80}{\scriptstyle \pm  0.001}$ & -\\
 \midrule 
\multirow{3}{*}{Covertype} & Time & $\mathbf{ 874.01}{\scriptstyle \pm  10.841}$ & $ 4073.99{\scriptstyle \pm  22.875}$ & - \\
 & ARI & $\mathbf{ 0.05}{\scriptstyle \pm  0.000}$ & $\mathbf{ 0.05}{\scriptstyle \pm  0.000}$ & - \\
 & NMI & $\mathbf{ 0.20}{\scriptstyle \pm  0.000}$ & $\mathbf{ 0.20}{\scriptstyle \pm  0.000}$ & -\\
\bottomrule 
 \end{tabular} 
 } 
 \caption{Experimental Results \label{tab:results}} 
 \end{table}

\begin{figure*}[t]
    \centering
    \begin{subfigure}{0.32\textwidth}
        \includegraphics[width=\linewidth]{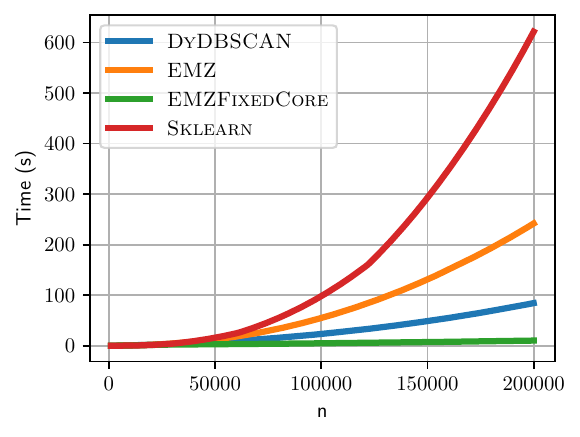}
        \caption{}
    \end{subfigure}
    \begin{subfigure}{0.32\textwidth}
        \includegraphics[width=\linewidth]{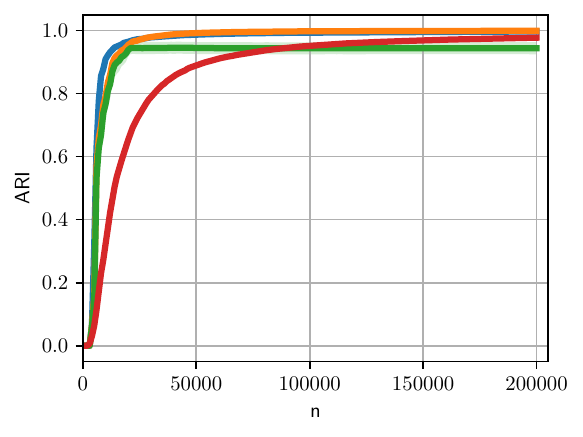}
        \caption{}
    \end{subfigure}
    \begin{subfigure}{0.32\textwidth}
        \includegraphics[width=\linewidth]{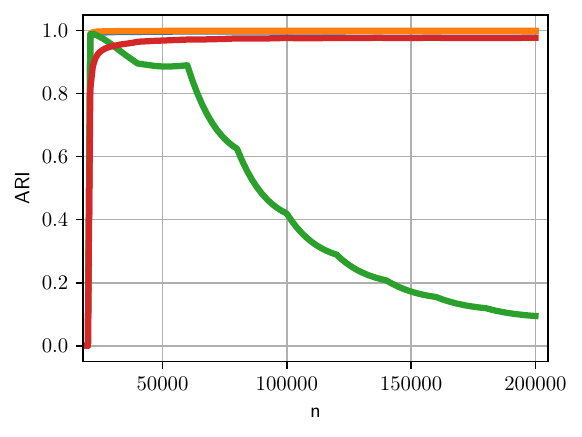}
        \caption{}
    \end{subfigure}
    \caption{\label{fig:blobs} Comparison on the blobs dataset. (a) The running time of each algorithm. (b) The ARI for each algorithm when data points are added in a random order. (c) The ARI for each algorithm when data points are added cluster-by-cluster.}
\end{figure*}

In this section, we evaluate the performance of our newly proposed dynamic DBSCAN algorithm by comparing it with several alternative methods.
All experiments are executed on a single thread using a 13th Gen Intel(R) Core(TM) i5-13500 processor.
We report the average results over 10 runs, with the standard error.
The code to reproduce the experiments is available at \url{https://github.com/seiyun-shin/dynamic_dbscan}.

\paragraph{Experimental setup.}
We conduct evaluations on a variety of real-world and synthetic datasets, with their properties summarized in~\Cref{tab:datasets}.
The blobs dataset is a synthetic dataset drawn from a mixture of Gaussians.
The other datasets are widely used for evaluating clustering and classification algorithms and are available on the OpenML dataset repository.
Detailed descriptions and license information for these datasets can be found on the OpenML website~\citep{OpenML2013}.

For the MNIST, Fashion-MNIST, and KDDCup99 datasets, we apply principal component analysis (PCA) to reduce the dimensionality to 20. For all datasets, we scale each dimension to have zero mean and unit variance.
In every experiment, we dynamically stream the data points to the algorithm in a random order, with a batch size of 1000.
After processing each batch, we compute the cluster labels for the entire dataset and evaluate the performance using two metrics: (1) the Adjusted Rand Index (ARI) and (2) Normalized Mutual Information (NMI).

\paragraph{Evaluation methods.}
To the best of our knowledge, our algorithm is the first dynamic version of DBSCAN, so we compare it with several simple baselines.
Specifically, the algorithms we evaluate are as follows:
\begin{enumerate}[nolistsep, topsep=0.2pt]
    \item \textsc{DynamicDBSCAN}: the dynamic DBSCAN algorithm proposed in this paper.
    \item \textsc{EMZ}: the near-linear time DBSCAN variant introduced by \citet{esfandiari2021almost}, where hash values for incoming points are computed once, and the graph is recomputed after processing each batch.
    \item \textsc{EMZFixedCore}: a variant of the \textsc{EMZ} algorithm which we propose and describe later.
    \item \textsc{Sklearn}: the standard DBSCAN implementation from the scikit-learn machine learning library~\citep{scikit-learn}.
\end{enumerate}
For both \textsc{DynamicDBSCAN} and \textsc{EMZ}, we set the hyperparameters $k = 10$ and $t = 10$ and $\epsilon = 0.75$.
We observe that the hyperparameters $k$ and $t$ are not sensitive, noting that their theoretical values of $O(\log (n))$ changes slowly with the number of data points. Hence, setting  $k$ and $t$ to their theoretical values would not significantly change the algorithm's performance.

\paragraph{Results and analysis.}
We report the results in~\Cref{tab:results}.
Due to the large memory requirement, we were unable to run the \textsc{Sklearn} algorithm on the KDDCup99 and Covertype datasets.
From these results, it is evident that the \textsc{DynamicDBSCAN} algorithm achieves a faster running time on dynamic datasets, with a similar performance when compared with the baseline algorithms.

\paragraph{Comparison with a fixed core point set.}
In addition to the \textsc{DynamicDBSCAN} algorithm, we propose another variant, \textsc{EMZFixedCore}, which builds upon the \textsc{EMZ} algorithm.
The \textsc{EMZFixedCore} algorithm processes the initial batch of data using the \textsc{EMZ} method, after which it keeps the set of core points fixed.
For subsequent updates, the algorithm treats each arriving point as a non-core point and assigns it to the same cluster as the first core point it collides with, determined by applying hash functions.

We conduct experiments on the blobs dataset under two scenarios.
In the first scenario, data points were added in a random order.
In the second, data points were added cluster-by-cluster, meaning that all points from cluster 1 were added before those from cluster 2, and so on.
We found that the running time of the algorithms is unaffected by the order of the incoming data points.
\Cref{fig:blobs} demonstrates that the \textsc{EMZFixedCore} algorithm exhibits a similar running time to \textsc{DynamicDBSCAN} and performs well when the order of the arriving data points is fully randomized.
On the other hand, when each cluster is added one at a time, the \textsc{EMZFixedCore} algorithm struggles to handle the increasing number of clusters, resulting in significantly poorer performance compared to \textsc{DynamicDBSCAN}.


%% file: sections_aistats/6_discussion.tex
\section{DISCUSSION} \label{sec:discussion}
To the best of our knowledge, we propose the first dynamic version of the DBSCAN, capable of efficiently handling evolving datasets with poly-logarithmic time complexity with respect to the total number of updates.

For future work, one intriguing direction would be to explore whether further reductions in the logarithmic dependency can be achieved.
Another promising direction is the development of a dynamic version of Hierarchical Density-Based Spatial Clustering of Applications with Noise (HDBSCAN)~\citep{campello2013density}.

%% file: sections_aistats/appendix.tex
\section{Proof of Lemma~\ref{lem:hash}}
Although the proof of~\Cref{lem:hash} is in~\citet{esfandiari2021almost}, we provide the full proof for completeness:
\begin{lemma}[\citep{esfandiari2021almost}]
    Given $\eps > 0$, the following holds for any two points $\vec{x}, \vec{y} \in \mathbb{R}^d$ and for a hash function $h$:
    \begin{enumerate}[nolistsep, topsep=0.2pt]
        \item $\Pr[h(\vec{x}) = h(\vec{y})] \geq 1 - \tfrac{\|\vec{x} - \vec{y}\|_1}{2\epsilon}$;
        \item $h(\vec{x}) = h(\vec{y}) \implies \|\vec{x} - \vec{y}\|_\infty \leq 2\epsilon$.
    \end{enumerate}
\end{lemma}

\begin{proof}
Fix a coordinate $j \in [d]$.
Observe that 
$\Pr[\lfloor \frac{\vec{x}_j + \eta}{2\epsilon} \rfloor \neq \lfloor \frac{y_j + \eta}{2\epsilon} \rfloor]$
is at most $\frac{|\vec{x}_j - y_j|}{2\epsilon}$.
By applying a union bound over all coordinates, we obtain $\Pr[h(\vec{x}) \neq h(\vec{y})] \leq \frac{\|\vec{x} - \vec{y}\|_1}{2\epsilon}$, proving the first part.

Now, suppose $\|\vec{x} - \vec{y}\|_\infty > 2\epsilon$.
Then there must exist a coordinate $j \in [d]$ such that $|\vec{x}_j - y_j| > 2\epsilon$.
This implies that $\lfloor \frac{\vec{x}_j + \eta}{2\epsilon}\rfloor \neq \lfloor \frac{y_j + \eta}{2\epsilon}\rfloor$ for any $\eta > 0$, which contradicts the assumption that $h(\vec{x}) = h(\vec{y})$.
Hence, we conclude that if $h(\vec{x}) = h(\vec{y})$, then $\|\vec{x} - \vec{y}\|_\infty \leq 2\epsilon$, proving the second part of the lemma.
This completes the proof of~\Cref{lem:hash}.
\end{proof}

\section{Deferred Details in~\ref{lem:non-core} via Order-wise Comparison} \label{app:lem3details}

In this section we present detailed derivations of the claim referenced in the proof of~\Cref{lem:non-core}.
Specifically, we aim to prove the following:
\begin{claim} \label{claim:order_compare}
    For any point $\vecx \in X$ with $\dist\left(\vecx, L_f(\lambda)\right) \geq 2 \left( \frac{\lambda}{R_1} \cdot 10 \frac{C_{\delta, n}}{\sqrt{k}} \right)^{\frac{1}{\beta}}$, it holds for sufficiently large $n$ that
    $\dist\left(\vecx, L_f(\lambda) \right) \geq 4 \sqrt{d} \varepsilon$.
\end{claim}
This claim is critical to the proof of~\Cref{lem:non-core}, as it shows that,
given $\dist(\vecy, L_f(\lambda)) \geq \dist(\vecx, L_f(\lambda)) - 2 \sqrt{d} \varepsilon$,
we can conclude $\dist(\vecy, L_f(\lambda)) \geq \tfrac{1}{2} \cdot \dist(\vecx, L_f(\lambda))$.
To prove the claim, we will consider two cases: (1) $d = \Theta(1)$ and (2) $d = \Theta(\log^c (n))$ for any constant $c > 0$.

\subsection{Proof of Claim~\ref{claim:order_compare} for \texorpdfstring{$d = \Theta(1)$}{d = O(1)}}
Taking $d$ to be a constant, we will show that
\begin{align*}
    \frac{\textsf{dist}(\vec{x}, L_f(\lambda))}{2} &= \left(\tfrac{1}{M_2}\right)^{\tfrac{1}{2\beta}}\cdot \omega\left((\log (n)/n)^{\tfrac{1}{2\beta + d}}\right) \mbox{ and }
    2 \sqrt{d} \epsilon = M_2^{\tfrac{1}{d}}\cdot O\left((\log (n)/n)^{\tfrac{1}{2\beta + d}}\right).
\end{align*}
Notice the parameter setup mentioned in~\Cref{subsubsec:assump_and_parameter} are:
\begin{itemize}[noitemsep, topsep=0.2pt]
    \item $k \geq M_1 \cdot \left(\tfrac{\lambda}{\lambda_c}\right)^2 \cdot \omega(\log (n)\cdot \log^{2+\alpha} \log (n))$ for some $\alpha > 0$;
    \item $k \leq M_2 \cdot \lambda^{\tfrac{2\beta + 2d}{2\beta + d}}\cdot O\left((\log (n))^{\tfrac{d}{2\beta + d}}\cdot n^{\tfrac{2\beta}{2\beta + d}}\cdot \log^{-\gamma} \log (n)\right)$ for some $\gamma > 0$;
    \item $\eps := \tfrac{1}{2}\left(\tfrac{k}{n\lambda(1-2C_{\delta, n}/\sqrt{k})}\right)^{\tfrac{1}{d}}$;
    \item $t := 100 \log \left(\tfrac{n}{\delta}\right)$;
\end{itemize}
where $C_{\delta, n} := C_0 \left(\log (t/\delta) \sqrt{d\log (n)}\right)$.

First notice that we have:
\begin{align*}
    \frac{\textsf{dist}(\vec{x}, L_f(\lambda))}{2} \geq \left( \frac{\lambda}{R_1} \cdot 10 \frac{C_{\delta, n}}{\sqrt{k}} \right)^{\frac{1}{\beta}}.
\end{align*}
Substituting the upper bound for \(k \leq M_2 \cdot \lambda^{\frac{2\beta + 2d}{2\beta + d}} (\log (n))^{\frac{d}{2\beta + d}}\cdot n^{\frac{2\beta}{2\beta + d}}\cdot \log^{-\gamma} \log (n)\),
we obtain:
\begin{align*}
    \frac{\textsf{dist}(\vec{x}, L_f(\lambda))}{2} \geq \left( \frac{10 C_0 \sqrt{d \log (n)} \log(t/\delta)\cdot \log^{\tfrac{\gamma}{2}} \log (n)}{R_1 \sqrt{M_2} \lambda^{\frac{\beta + d}{2\beta + d}} (\log (n))^{\frac{d}{2(2\beta + d)}} n^{\frac{\beta}{2\beta + d}}} \right)^{\frac{1}{\beta}}.
\end{align*}
Notice that the exponent of $\log (n)$ simplifies to:
\begin{align*}
    \left(\frac{1}{2} - \frac{d}{2(2\beta+d)}\right)\cdot \frac{1}{\beta} = \left(\frac{2\beta+d - d}{2(2\beta+d)}\right) \cdot \frac{1}{\beta} =  \frac{1}{2\beta+d},
\end{align*}
which will lead us to obtain the order of \(\frac{\textsf{dist}(\vec{x}, L_f(\lambda))}{2}\) as:
\begin{align*}
    \left(\frac{1}{M_2}\right)^{\frac{1}{2\beta}}\cdot \omega\left(\left(\frac{\log (n)}{n}\right)^{\frac{1}{2\beta + d}}\cdot \log^{\frac{\gamma}{2\beta}} \log (n)\right).
\end{align*}

For \(2\sqrt{d} \epsilon\), we have:
\begin{align*}
    2\sqrt{d}\epsilon = \sqrt{d} \left( \frac{k}{n\lambda(1 - 2C_{\delta, n}/\sqrt{k})} \right)^{\frac{1}{d}}.
\end{align*}
Substituting the upper bound for \(k\), we obtain:
\begin{align*}
    2\sqrt{d}\epsilon \leq \sqrt{d} \left( \frac{M_2 \lambda^{\frac{2\beta + 2d}{2\beta + d}} (\log (n))^{\frac{d}{2\beta + d}} n^{\frac{2\beta}{2\beta + d}}\cdot \log^{-\gamma} \log (n)}{n \lambda(1 - 2C_{\delta, n}/\sqrt{k})} \right)^{\frac{1}{d}}.
\end{align*}
Observe that using the lower bound for $k$, we can estimate:
\begin{align*}
    \frac{C_{\delta, n}}{\sqrt{k}}
    \leq \frac{C_0 \left(\log (t/\delta) \sqrt{d\log (n)}\right)}{\sqrt{\log (n) \cdot \log^{2+\alpha} \log (n)}}
    = O\left(\frac{1}{\log^{\frac{\alpha}{2}}\log (n)}\right) \rightarrow 0,\quad \text{as}\quad n \rightarrow \infty.
\end{align*}
Hence approximating for small \(\frac{C_{\delta, n}}{\sqrt{k}}\) for sufficiently large $n$, we simplify:
\begin{align*}
    2\sqrt{d} \epsilon
    \leq \sqrt{d} \left( \frac{M_2 \lambda^{\frac{2\beta + 2d}{2\beta + d} - 1} (\log (n))^{\frac{d}{2\beta + d}} n^{\frac{2\beta}{2\beta + d}}\cdot \log^{-\gamma} \log (n)}{n} \right)^{\frac{1}{d}}.
\end{align*}
As the exponent of $n$ becomes $\left(\tfrac{2\beta}{2\beta+d} - 1\right)\cdot \tfrac{1}{d} = -\tfrac{1}{2\beta+d}$,
the order of \(2 \sqrt{d} \epsilon\) is then:
\begin{align*}
M_2^{\frac{1}{d}}\cdot O\left(\left(\frac{\log (n)}{n}\right)^{\frac{1}{2\beta + d}}\cdot \log^{-\frac{\gamma}{d}} \log (n)\right).
\end{align*}

In conclusion, both \(\frac{\textsf{dist}(\vec{x}, L_f(\lambda))}{2}\) and \(2\sqrt{d} \epsilon\) have the same asymptotic behaviour with respect to $n$, which is \(\Theta\left(\left(\frac{\log (n)}{n}\right)^{\frac{1}{2\beta + d}}\right)\); however, the constant factor for \(2\sqrt{d} \epsilon\) is dependent on \(M_2^{\frac{1}{d}}\), while for \(\frac{\textsf{dist}(\vec{x}, L_f(\lambda))}{2}\), it involves \(\left(\frac{1}{M_2}\right)^{\frac{1}{2\beta}}\).
Consequently, if we choose $M_2$ sufficiently small and $n$ large enough, we can ensure \(\frac{\textsf{dist}(\vec{x}, L_f(\lambda))}{2} > 2\sqrt{d} \epsilon\).

This completes the proof. \hfill $\qed$

\subsection{Proof of Claim~\ref{claim:order_compare} for \texorpdfstring{$d = \Theta(\log^c (n))$}{d = O(log\^c(n))} and Constant \texorpdfstring{$c > 0$}{c > 0}}
In this subsection, we analyse the order-wise comparison between $\frac{\textsf{dist}(\vec{x}, L_f(\lambda))}{2}$ and $2 \sqrt{d} \epsilon$ with respect to $d$ and $n$.
To account for the dependence on $d$, we use the following detailed parameter setup:
\begin{itemize}[noitemsep, topsep=0.2pt]
    \item $100 \cdot (100 \sqrt{d})^{2\beta + d} \left( \tfrac{\lambda}{\lambda_c} \right)^2 \left( \tfrac{R_2}{R_1} \right)^2 \cdot C_{\delta, n}^2 \sqrt{d \log (n)} \cdot \log^{2+\alpha} \log (n) \leq k \leq \left( \tfrac{C_{\delta,n}}{R_2} \right)^{\tfrac{2d}{2\beta + d}} \left( \tfrac{1}{4d} \right)^{\tfrac{\beta d}{2\beta + d}} \lambda^{\tfrac{2\beta + 2d}{2\beta + d}} n^{\tfrac{2\beta}{2\beta + d}}\cdot \log^{-\gamma} \log (n)$ for some $\alpha, \gamma > 0$;
    \item $\eps := \tfrac{1}{2}\left(\tfrac{k}{n\lambda(1-2C_{\delta, n}/\sqrt{k})}\right)^{\tfrac{1}{d}}$; and
    \item $t := 100 \log \left(\tfrac{n}{\delta}\right)$;
\end{itemize}
where $C_{\delta, n} = C_0 \left(\log (t/\delta) \sqrt{d\log (n)}\right)$.
Using the upper bound for $k$, we now substitute into the expression for $\frac{\textsf{dist}(\vec{x}, L_f(\lambda))}{2}$, yielding:
\begin{align*}
    \frac{\textsf{dist}(\vec{x}, L_f(\lambda))}{2}
    \geq \left( \frac{\lambda}{R_1} \cdot 10 \frac{C_{\delta, n}}{\sqrt{k}} \right)^{\frac{1}{\beta}}
    \geq \left( \frac{10}{R_1} \cdot \frac{C_0 \log(t/\delta) \sqrt{d \log (n)}\cdot \log^{\tfrac{\gamma}{2}} \log (n)}{\left( \frac{C_0 \log(t/\delta) (d \log (n))^{1/2}}{R_2} \right)^{\frac{d}{2\beta + d}} \cdot \left( \frac{1}{4d} \right)^{\frac{\beta d}{2(2\beta + d)}} \lambda^{\frac{\beta + d}{2\beta + d}} n^{\frac{\beta}{2\beta + d}}} \right)^{\frac{1}{\beta}}.
\end{align*}
After simplifying the terms involving \(d\), \(\log \log (n)\), \(\log (n)\), and \(n\), the expression becomes:
\begin{align*}
    \frac{\textsf{dist}(\vec{x}, L_f(\lambda))}{2} = \omega\left(d^{\frac{2+d}{2(2\beta + d)}} \cdot \left(\frac{\log (n)}{n}\right)^{\frac{1}{2\beta + d}}\cdot \log^{\frac{\gamma}{2\beta}} \log (n) \right),
\end{align*}
where the exponent for $\tfrac{\log (n)}{n}$ and $\log \log (n)$ follow from the same analysis as for $d = \Theta(1)$, and the the exponent for $d$ comes from:
\begin{align*}
    \left(\frac{1}{2} - \frac{d}{2(2\beta+d)} + \frac{\beta d}{2(2\beta + d)}\right)\cdot \frac{1}{\beta} = \left(\frac{2\beta+d - d + \beta d}{2(2\beta+d)}\right) \cdot \frac{1}{\beta} =  \frac{2+d}{2(2\beta+d)}.
\end{align*}
Next, using the approximation \(1 - 2C_{\delta, n}/\sqrt{k} \approx 1\) for sufficiently large $n$, we simplify:
\(
2\sqrt{d}\epsilon \approx \sqrt{d} \left( \frac{k}{n\lambda} \right)^{\frac{1}{d}}.
\)
Substituting the upper bound for $k$ into the expression for $\eps$, we obtain:
\begin{align*}
    2\sqrt{d}\epsilon
    = \sqrt{d}\left(\left(\frac{1}{n \lambda}\right) \cdot \left( \frac{C_0 \log(1/\delta) \sqrt{d \log (n)}}{R_2} \right)^{\frac{2d}{2\beta + d}} \cdot \left( \frac{1}{4d} \right)^{\frac{\beta d}{2\beta + d}} \cdot \lambda^{\frac{2\beta + 2d}{2\beta + d}}\cdot n^{\frac{2\beta}{2\beta + d}}\cdot \log^{-\gamma}\log (n)\right)^{\frac{1}{d}}.
\end{align*}
Simplifying further, we have:
\begin{align*} 
    2\sqrt{d}\epsilon = O\left(d^{\frac{2 + d}{2(2\beta + d)}}\cdot \left(\frac{\log (n)}{n}\right)^{\frac{1}{2\beta+d}}\cdot \log^{-\frac{\gamma}{d}} \log (n)\right),
\end{align*}
where the exponent of $d$ arises from:
\begin{align*}
    \frac{1}{2} + \left(\frac{1}{2}\cdot \frac{2d}{2\beta + d} -\frac{\beta d}{2\beta + d}\right)\cdot \frac{1}{d} = \frac{1}{2} + \frac{1 - \beta}{2\beta + d} = \frac{2\beta + d + 2(1 - \beta)}{2(2\beta + d)} = \frac{2 + d}{2(2\beta + d)}.
\end{align*}
Therefore, both \(\frac{\textsf{dist}(\vec{x}, L_f(\lambda))}{2}\) and \(2\sqrt{d} \epsilon\) exhibit the same asymptotic behaviour with respect to $d$ and $n$, namely:
\begin{align*}
    d^{\frac{2 + d}{2(2\beta + d)}}\cdot \left(\frac{\log (n)}{n}\right)^{\frac{1}{2\beta+d}},
\end{align*}
except for the $\log \log n$ factor.
The presence of the $\log \log n$ dependency implies that for sufficiently large $n$, we can ensure \(\frac{\textsf{dist}(\vec{x}, L_f(\lambda))}{2} > 2\sqrt{d} \epsilon\).

Lastly, we highlight that the regime of $d = \Theta(\log^c (n))$ for a constant $c > 0$ is essential for ensuring that the volume of the cubes with side-length proportional to $\varepsilon$ used in density level estimation remains bounded, even when $n$ grows large.
This completes the proof. \hfill $\qed$